\RequirePackage{amsmath} 
\documentclass[runningheads]{llncs}

\usepackage{color}
\usepackage[width=122mm,left=12mm,paperwidth=146mm,height=193mm,top=12mm,paperheight=217mm]{geometry}

\usepackage{adjustbox}
\usepackage{amsmath}
\usepackage{tabularx} 
\usepackage{scalerel} 
\usepackage[numbers,sort&compress,square]{natbib} 

\newcommand{\minus}{\scalebox{0.75}[1.0]{$-$}} 
\DeclareMathOperator*{\argmax}{argmax}
\DeclareMathOperator*{\softmax}{softmax}
\newcommand{\primecus}{\hstretch{0.7}{\vstretch{0.7}{^\prime}}}
\newcolumntype{C}[1]{>{\centering\let\newline\\\arraybackslash\hspace{0pt}}m{#1}} 
\newcolumntype{L}[1]{>{\let\newline\\\arraybackslash\hspace{0pt}}m{#1}} 
\makeatletter
\newcommand\mypartop{\@startsection{paragraph}{4}{\z@}
                                    {3.25ex \@plus1ex \@minus.2ex}	
                                    {-1em}%
                                    {\bfseries \normalsize}}
\newcommand\mypar{\@startsection{paragraph}{4}{\z@}
                                    {1.00ex \@plus1ex \@minus.2ex}	
                                    {-1em}%
                                    {\bfseries \normalsize}}
\renewcommand\@biblabel[1]{#1.} 
\makeatother

\begin{document}
\pagestyle{headings}
\mainmatter

\title{Region-based semantic segmentation \\ with end-to-end training}

\titlerunning{Region-based semantic segmentation with end-to-end training}

\authorrunning{Caesar et al.}

\author{Holger Caesar, Jasper Uijlings, Vittorio Ferrari}
\institute{University of Edinburgh}

\maketitle

\begin{abstract}
We propose a novel method for semantic segmentation, the task of labeling each pixel in an image with a semantic class.
Our method combines the advantages of the two main competing paradigms.
Methods based on region classification offer proper spatial support for appearance measurements,
but typically operate in two separate stages, none of which targets pixel labeling performance at the end of the pipeline.
More recent fully convolutional methods are capable of end-to-end training for the final pixel labeling, but resort to fixed patches as spatial support.
We show how to modify modern region-based approaches to enable end-to-end training for semantic segmentation.
This is achieved via a differentiable region-to-pixel layer and a differentiable free-form Region-of-Interest pooling layer.
Our method improves the state-of-the-art in terms of class-average accuracy with $64.0\%$ on SIFT Flow and $49.9\%$ on PASCAL Context, and is particularly accurate at object boundaries.
\end{abstract}


\section{Introduction}

We address the task of semantic segmentation, labeling each pixel in an image with a semantic class.
Currently, there are two main paradigms: classical region-based
approaches~\cite{boix12ijcv,caesar15bmvc,carreira12eccv,dai15cvpr,george15cvpr,girshick14cvpr,hariharan14eccv,li13cvpr,mostajabi15cvpr,plath09icml,sharma14nips,sharma15cvpr,tighe10eccv,tighe13cvpr,tighe14cvpr,yang14cvpr,mottaghi14cvpr}
and, inspired by the Convolutional Neural Network~(CNN) revolution, fully convolutional
approaches~\cite{chen15iclr,dai15iccv,eigen15iccv,farabet13pami,hariharan15cvpr,long15cvpr,noh15iccv,pinheiro14icml,zheng15iccv}.

\begin{figure}[t]
\centering
\includegraphics[width=1.0\textwidth]{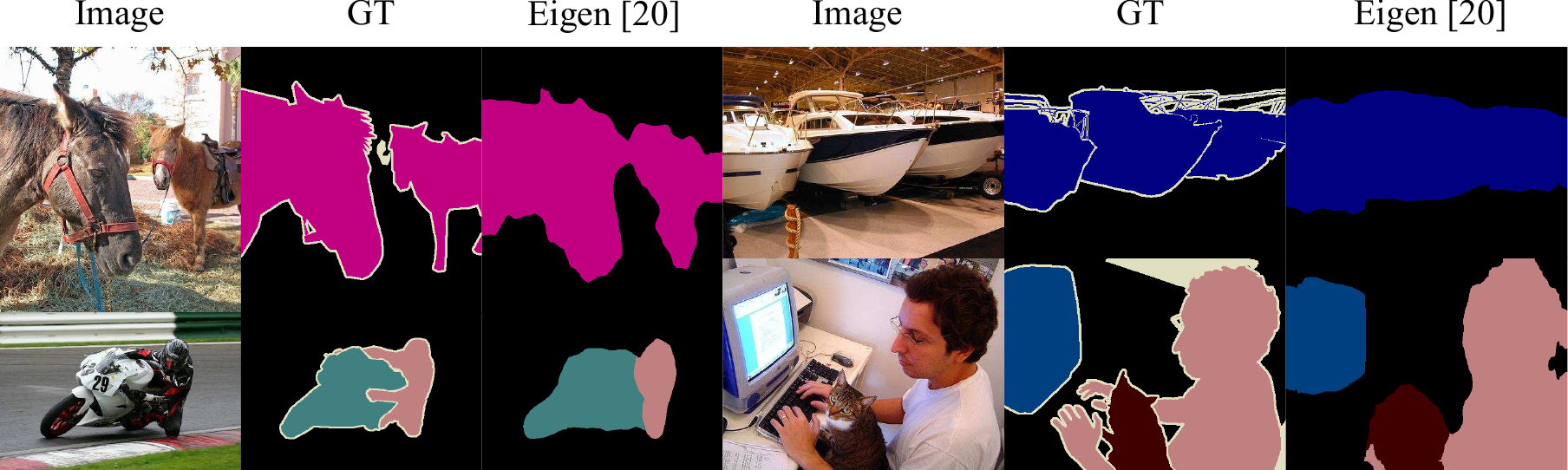}
\caption{\small 
\textit{
Fully convolutional methods typically produce fuzzy object boundaries, as illustrated here by examples from Eigen and Fergus~\cite{eigen15iccv}.
}
}
\label{fig:boundary-comparison}
\end{figure}

In the fully convolutional approach the idea is to directly learn a mapping from image pixels to
class labels using a CNN.  This results in a single model, directly
optimized end-to-end for the task at hand, including the intermediate image representations (i.e. the
hidden layers in the network).  However, the spatial support on which predictions are based are
fixed-size square patches of the input image. Intuitively, this is suboptimal since: (I) Objects are
free-form rather than square, so ideally the intermediate representations should take this into
account. (II) Objects do not have a fixed size, but occur at various scales. Hence
many patches either cover pieces of multiple objects and mix their representations, or cover a piece of an
object, which is sometimes difficult to recognize in isolation
(e.g. a patch on the belly of a cow).
An additional problem is that fully convolutional methods typically make predictions at a coarse resolution, which often results in inaccurate object boundaries~\cite{chen15iclr,eigen15iccv,farabet13pami,hariharan15cvpr,noh15iccv,zheng15iccv}.
Fig.~\ref{fig:boundary-comparison} illustrates this on example outputs of~\cite{eigen15iccv}.

In the region-based approach, the image is first segmented into coherent regions, which are
described by image
features~\cite{boix12ijcv,caesar15bmvc,carreira12eccv,dai15cvpr,george15cvpr,girshick14cvpr,hariharan14eccv,li13cvpr,mostajabi15cvpr,plath09icml,sharma14nips,sharma15cvpr,tighe10eccv,tighe13cvpr,tighe14cvpr,yang14cvpr}.
Typically many regions are extracted at multiple
scales~\cite{caesar15bmvc,carreira12eccv,dai15cvpr,girshick14cvpr,hariharan14eccv,li13cvpr,plath09icml,sharma14nips,sharma15cvpr},
capturing complete objects and canonical object parts (e.g. faces) which in turn facilitates recognition. Furthermore, the segmentation process delivers regions which
follow object boundaries quite well.
However, these methods generally first extract region features and then train a classifier optimized for classifying regions rather than for the final semantic segmentation criterion (i.e. pixel-level labeling)~\cite{caesar15bmvc,carreira12eccv,dai15cvpr,girshick14cvpr,hariharan14eccv,li13cvpr,plath09icml}.
Hence, while these methods benefit from the power of multi-scale, overlapping regions, they cannot be trained
end-to-end for semantic segmentation.

In this paper we want the best of both worlds. We propose a region-based semantic segmentation model with an accompanying end-to-end training scheme based on a CNN architecture (Fig.~\ref{fig:cnn-architectures}c). To enable this we introduce a novel, differentiable region-to-pixel
layer which maps from regions to image pixels. We insert this layer before the final classification
layer, enabling the use of a pixel-level loss which allows us to directly optimize for semantic
segmentation.  Conceptually, our region-to-pixel layer ignores regions which have low activations
for all classes and which therefore do not impact the final labeling. This is in contrast to
all multi-scale region-based methods where such regions incorrectly affect
training~\cite{caesar15bmvc,carreira12eccv,dai15cvpr,girshick14cvpr,hariharan14eccv,li13cvpr,plath09icml}.
Additionally, we introduce a differentiable Region-of-Interest pooling layer which
operates on the final convolutional layer in the spirit of Fast R-CNN~\cite{girshick15iccv}, but
which is adapted for free-form regions like~\cite{dai15cvpr,sharma14nips,sharma15cvpr}.
Note how we use region proposals from a separate pre-processing stage. By end-to-end we mean training all parameters for the final pixel-level loss, rather than for region classification.

To summarize, our contributions are:
(1) We introduce a region-to-pixel layer which enables full end-to-end training of semantic segmentation models based on multi-scale overlapping regions.
(2) We introduce a Region-of-Interest pooling layer specialized for free-form regions.
(3) We obtain state-of-the-art results on the SIFT Flow and the PASCAL Context datasets, in terms of class-average accuracy.
Our approach delivers crisp object boundaries, as demonstrated in Fig.~\ref{fig:examples-pascalcontext} and Sect.~\ref{sec:extra-analysis}.
We release the source code of our method at https://github.com/nightrome/matconvnet-calvin

\section{Related Work}
\label{sec:relatedwork}

\subsection{Region-based semantic segmentation}

Region-based semantic segmentation methods first extract free-form regions~\cite{carreira10cvpr,uijlings13ijcv,endres14pami,arbelaez14cvpr} from an image and describe them with features.
Afterwards a region classifier is trained. At test time, region-based predictions are mapped to pixels, usually by labeling a pixel according to the highest scoring region that contains it.
Region-based methods generally yield crisp object boundaries~\cite{boix12ijcv,caesar15bmvc,carreira12eccv,dai15cvpr,george15cvpr,girshick14cvpr,hariharan14eccv,li13cvpr,mostajabi15cvpr,plath09icml,sharma14nips,sharma15cvpr,tighe10eccv,tighe13cvpr,tighe14cvpr,yang14cvpr,mottaghi14cvpr}.
Fig.~\ref{fig:cnn-architectures}b shows a prototypical architecture for such an approach (which we modernized by basing it on Fast R-CNN~\cite{girshick15iccv}). We discuss several aspects below.

\mypar{Multi-scale vs single-scale regions.}
Several region-based methods use an oversegmentation to create small, non-overlapping
regions~\cite{boix12ijcv,george15cvpr,mostajabi15cvpr,tighe10eccv,tighe13cvpr,tighe14cvpr,yang14cvpr}.
Intuitively however, objects are more easily recognized as a whole than by looking at small object
parts individually. The inherent multi-scale aspect of recognition is adequately captured in many recent works
using multi-scale, overlapping regions~\cite{caesar15bmvc,carreira12eccv,dai15cvpr,girshick14cvpr,hariharan14eccv,li13cvpr,plath09icml,sharma14nips,sharma15cvpr}.

\mypar{Training criterion.}
The final criterion is pixel-level prediction of class labels. However,
we use overlapping regions whose predictions are in competition with each other on the pixel level.
Typically, many methods initially ignore this by simply training a classifier to predict region labels~\cite{caesar15bmvc,carreira12eccv,dai15cvpr,girshick14cvpr,hariharan14eccv,li13cvpr,plath09icml},
which is \emph{different} from semantic segmentation (Sec.~\ref{sec:regionbased}).
At test time one labels a pixel by simply taking the maximum over all regions containing it~\cite{caesar15bmvc,carreira12eccv,dai15cvpr,girshick14cvpr,hariharan14eccv}.
A few works partially addressed the mismatch between training and test time through a post-processing stage using graphical models~\cite{li13cvpr,plath09icml} or by joint calibration~\cite{caesar15bmvc}. However, none of them does full end-to-end training.

\mypar{Region representations.}
Most older works use hand-crafted region-based features~\cite{boix12ijcv,carreira12eccv,george15cvpr,li13cvpr,plath09icml,tighe10eccv,tighe13cvpr,tighe14cvpr,yang14cvpr}
often based on~\cite{carreira12eccv,tighe10eccv}.
More recent works instead use the top convolutional layers of a pre-trained CNN
(e.g.~\cite{krizhevsky12nips,simonyan15iclr}) as feature
representations~\cite{caesar15bmvc,dai15cvpr,girshick14cvpr,hariharan14eccv,mostajabi15cvpr,sharma14nips,sharma15cvpr}.
These representations can be free-form respecting the
shape of the region~\cite{dai15cvpr,girshick14cvpr,hariharan14eccv,mostajabi15cvpr,sharma14nips,sharma15cvpr} or
simply represent the bounding box around the region~\cite{caesar15bmvc,girshick14cvpr}. Furthermore
regions can be cropped out from the image before being fed to the
network~\cite{girshick14cvpr,hariharan14eccv,mostajabi15cvpr} or one can create region
representations from a convolutional layer~\cite{dai15cvpr,sharma14nips,sharma15cvpr}, termed
Region-of-Interest~(ROI) pooling~\cite{girshick15iccv} or Convolutional Feature Masking~\cite{dai15cvpr}.
CNN representations become more powerful when further trained for the task.
In~\cite{caesar15bmvc,girshick14cvpr,hariharan14eccv} they train CNNs, but for the task of region
classification, not for semantic segmentation.

\subsection{Fully convolutional semantic segmentation}

Fully convolutional methods learn a direct mapping from pixels to pixels, which was pioneered
by~\cite{shotton09ijcv} in the pre-CNN era. Early CNN-based approaches train relatively shallow
end-to-end networks~\cite{farabet13pami,pinheiro14icml}, whereas more recent works use much deeper
networks whose weights are initialized by pre-training on the ILSVRC~\cite{russakovsky15ijcv}
image classification
task~\cite{chen15iclr,dai15iccv,eigen15iccv,hariharan15cvpr,long15cvpr,noh15iccv,zheng15iccv}. The
main insight to adapt these networks for semantic segmentation was to re-interpret the
classification layer as 1x1 convolutions~\cite{sermanet14iclr,long15cvpr}. A prototypical model is illustrated in
Fig.~\ref{fig:cnn-architectures}a.

\mypar{Square receptive fields.}
All fully convolutional methods have receptive fields of fixed
shape
(square)~\cite{chen15iclr,dai15iccv,eigen15iccv,farabet13pami,hariharan15cvpr,long15cvpr,noh15iccv,pinheiro14icml,zheng15iccv}.
However, since objects are free-form this may be suboptimal. 

\mypar{Multi-scale.} Recognition is a multi-scale problem, which is addressed by using two
strategies:
\emph{(I) Multi-scale representations.} Using skip-layer connections~\cite{bishop95book,ripley96book}, representations from different convolutional layers can be combined~\cite{eigen15iccv,hariharan15cvpr,long15cvpr,pinheiro14icml}.
This leads to multi-scale representations of a predetermined size.
\emph{(II)~Multi-scale application.} In~\cite{hariharan15cvpr,noh15iccv} they train and apply
their method on multi-scale, rectangular image crops. However, this results in a mismatch
between training time, where each crop is considered separately, and test time,
where predictions of multiple crops are combined before evaluation.

\mypar{Fuzzy object boundaries.}
It is widely acknowledged that fully convolutional approaches yield rather fuzzy object
boundaries~\cite{chen15iclr,eigen15iccv,farabet13pami,hariharan15cvpr,noh15iccv,zheng15iccv}. A
variety of strategies address this.
\emph{(I) Multi-scale.} The multi-scale methods discussed
above~\cite{eigen15iccv,hariharan15cvpr,long15cvpr,noh15iccv,pinheiro14icml} include a fine scale
resulting in improved object boundaries.
\emph{(II) Conditional Random Fields (CRFs).} CRFs are a
classical tool to refine pixel-wise labelings and are used as post-processing step
by~\cite{chen15iclr,farabet13pami,noh15iccv,zheng15iccv}.  Notably,~\cite{zheng15iccv} reformulate
the CRF as a recurrent neural network enabling them to train the whole network including
convolutional layers in an end-to-end fashion.
\emph{(III) Post-processing by region proposals.}
Finally,~\cite{farabet13pami} averages pixel-wise network outputs over regions from an oversegmentation.

\subsection{This paper}
We propose a model based on free-form, multi-scale, overlapping regions. We design a partially
differentiable region-to-pixel layer enabling end-to-end training for semantic segmentation.
Additionally we introduce a ROI pooling layer which is
free-form~\cite{dai15cvpr,sharma14nips,sharma15cvpr} yet
also differentiable~\cite{girshick15iccv}.
\section{Method}

\begin{figure}[t]
\begin{center}
\includegraphics[width=1.0\textwidth]{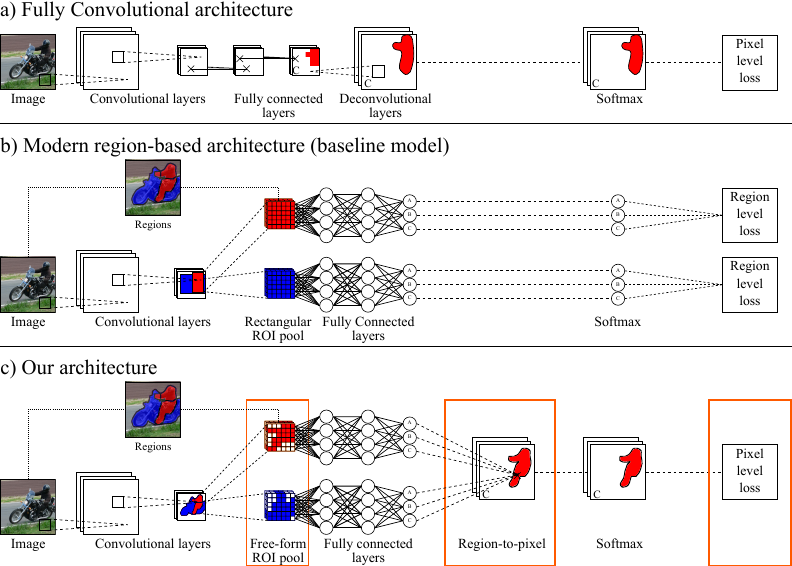}
\end{center}
\caption{
\textit{
\small
Overview of three semantic segmentation architectures.
We show only layers with trainable parameters, softmax and loss layers. We omit all pre- and post-processing steps.
a) shows the class of fully convolutional architectures that are end-to-end trainable, but do not have regions.
b) shows the baseline model, representative for modern region-based architectures.
It is not end-to-end trainable for the desired pixel labeling criterion.
c) shows our suggested architecture, which pools activations of each region in a free-form manner,
maps the region-level predictions to pixels and computes a loss at the pixel level.
Hence our method combines regions and end-to-end training.
Our main contributions are highlighted by orange boxes.
}
}
\label{fig:cnn-architectures}
\end{figure}

Section~\ref{sec:regionbased} presents a baseline model that is representative for modern region-based semantic segmentation~\cite{caesar15bmvc,dai15cvpr,girshick14cvpr,hariharan14eccv} (Fig.~\ref{fig:cnn-architectures}b),
and explains its shortcomings.
Sections~\ref{sec:endtoend}-\ref{sec:efficienteval} present our framework, which addresses these issues (Fig.~\ref{fig:cnn-architectures}c).

\subsection{Region-based semantic segmentation}
\label{sec:regionbased}

\mypartop{Model.} Fig.~\ref{fig:cnn-architectures}b presents a typical region-based semantic
segmentation architecture. It
modernizes~\cite{caesar15bmvc,dai15cvpr,girshick14cvpr,hariharan14eccv} by using the Region-of-Interest pooling layer
of~\cite{girshick15iccv}. We use this model as a baseline in our experiments
(Sec.~\ref{sec:experiments}).

The input to the network are images and free-form regions~\cite{uijlings13ijcv}.  The image is fed
through several convolutional layers. A Region-of-Interest pooling layer~\cite{girshick15iccv}
creates a feature representation of the tight bounding boxes around each region. These region features
are then fed through several fully connected layers and a classification layer, followed by a softmax,
resulting in region-level predictions. At test time, these predictions are mapped from regions to
pixels: each pixel $p$ is assigned the label $o_p$ with the highest probability over all classes and
all regions containing $p$:
\begin{equation}
    o_p = \argmax\limits_c \, \max_{r \ni p} \, \softmax_c S_{r,c}
    \label{eq:model-simple}
\end{equation}
Here $S_{r,c}$ denotes the classifier scores for region
$r$ and class $c$ (i.e.  activations of the classification layer).

\mypar{Training.}
The training procedure searches for the network parameters that minimize a cross-entropy log-loss $\mathcal{L}$ over regions:
\begin{equation}
\mathcal{L} = \minus \sum_c \frac{1}{R} \sum_{r=1}^{R} \; y_{r,c} \; \log{ \, \softmax_c \, S_{r,c}}
\label{eq:loss-simple}
\end{equation}
Here $R$ indicates the number of regions in the training set and $y_{r,c} \in \{0, 1\}$ is a ground truth label indicating whether region $r$ has label $c$.
The network is trained with Stochastic Gradient Descent~(SGD) with momentum.
To update the network weights, one needs to compute the partial derivatives of the loss with respect to the weights.
These derivatives depend on the partial derivatives of the loss with respect to the outputs of the respective layer.

\mypar{Problems.}
A first problem arises because the softmax is applied before pixel assignment in Eq.~\eqref{eq:model-simple}:
(I) regions with low but highly varying activation scores are unsure about the class, but
can still yield high probabilities due to the softmax. Intuitively, this means that such
non-discriminative regions can wrongly affect the final prediction.

More importantly, since $\max_{r \ni p}$ occurs at test time (Eq.~\eqref{eq:model-simple}),
but not at training time (Eq.~\eqref{eq:loss-simple}),
the pixel-wise evaluation criterion at test time is \emph{different} from
the region-level optimization criterion at training time. This has several consequences:
(II) While during training \emph{all} regions affect the network, at test time most regions are ignored.
(III) It is unclear what are good region training examples for achieving good performance at test time: Are
positive examples only ground truth regions? Or should we use also region proposals which partially overlap
with the ground truth? And with what threshold? What overlap are negative proposals allowed to have
to count as negative examples? Hence one has to select overlap thresholds for positive and negative
examples empirically using test time evaluations.
(IV) Regions with different size have the same weight.
(V) The network is not trained end-to-end for semantic segmentation, but for the intermediate task of region classification instead. Hence both the classification layer and the representation layers will be suboptimal for the actual semantic segmentation task.

\subsection{End-to-end training for region-based semantic segmentation}
\label{sec:endtoend}

\mypartop{Model.}
To combine the paradigms of region-based semantic segmentation and end-to-end training,
we map from regions to pixels as in Eq.~\eqref{eq:model-simple}, but \textit{before} the softmax and loss computation on a pixel-level:
\begin{equation}
o_p = \argmax\limits_c \, \softmax_c \, \max_{r \ni p} \, S_{r,c}
\label{eq:model-advanced}
\end{equation}
This region-to-pixel layer is shown in Fig.~\ref{fig:cnn-architectures}c. It brings two benefits.
At training time, having the region-to-pixel layer before the loss enables optimizing a pixel-level loss.
Furthermore, having the region-to-pixel layer before the softmax ensures that the class score for each pixel is taken from the region with the
highest activation score, hence each class can be recognized at its appropriate scale.

\mypar{Training.}
In Eq.~\eqref{eq:loss-simple} the baseline model computes a cross-entropy log-loss on the region-level.
Here instead we compute a log-loss on the pixel-level:
\begin{equation}
\mathcal{L} = \minus \sum_c \frac{1}{P} \sum_{p=1}^{P} \; y_{p,c} \; \log{\softmax_c S_{p,c}}
\label{eq:loss-advanced}
\end{equation}
Here $P$ indicates the number of pixels in the training set, $y_{p,c} \in \{0, 1\}$ indicates whether pixel p has ground truth label $c$, and $S_{p,c} = \max_{r \ni p} \, S_{r,c}$ is the pixel-level score for
class $c$.
As in Section~\ref{sec:regionbased} we train the network using SGD.
To determine the partial derivatives of our region-to-pixel layer, we observe that
it does not have any weights and we only need to compute the subgradients of the
loss with respect to the region-level scores $S_{r,c}$:
\begin{equation}
\frac{\partial \mathcal{L}}{\partial S_{r,c}} =
\sum\limits_{
p \, \in \, r \;   | \; r \, = \, \argmax_{r\primecus \ni p\primecus} \, S_{r\primecus,c}}
\frac{\partial \mathcal{L}}{\partial S_{p,c}}
\end{equation}
This means that for each class we map each pixel-level gradient to the region with the highest score among all regions that include the pixel.
If multiple pixels per class map to the same region, their gradient
contributions are summed.

\mypar{Advantages.}
Our model addresses all problems raised in Sec.~\ref{sec:regionbased}: 
(I) Pixels are always labeled according to the relevant region with the highest activation score for that class.
(II) Regions which do not affect the pixel-level prediction are ignored during training.
(III) Since we evaluate pixels there is no need to assign class labels to regions for training.
(IV) The pixel-level loss is agnostic to different sizes of region proposals.
(V) We train our method end-to-end for the actual semantic segmentation criterion, resulting in properly optimized classifiers and region representations.

\subsection{Pooling on free-form regions}
\label{sec:freeform}

\mypartop{Model.}
While the baseline model classifies free-form regions, their feature representations are computed on the bounding box. This is suboptimal as the regions can take highly irregular shapes. We propose here a free-form Region-of-Interest~(ROI) pooling layer which computes representations taking into account only pixels actually in the region (Fig.~\ref{fig:cnn-architectures}c):
\begin{equation}
S_{i, d, r}^R = 
\max\limits_{
j \; | \; \phi(j) \, = \, i , \; \delta_{j, r} \, = \, 1
}
\; S_{j, d}^C
\end{equation}
Here $S_{i, d, r}^R$ is the ROI pooling activation for ROI coordinate $i$, channel $d$ and region $r$.
For each ROI coordinate and channel we maximize over the corresponding coordinate $j$ in the convolutional map $S_{j, d}^C$,
considering only points inside the region, i.e. $\delta_{j, r} = 1$.
The mapping $\phi$ from convolutional map coordinates to ROI ones is done as in~\cite{girshick15iccv,he14eccv},
but operates on a free-form region rather than a bounding box.

\mypar{Training.}
During the forward pass the highest scoring convolutional map coordinate $\pi(i, d, r)$ for each ROI coordinate and channel is computed as:
\begin{equation}
\pi(i, d, r) = 
\argmax\limits_{
j \; | \; \phi(j) \, = \, i , \; \delta_{j, r} \, = \, 1
}
\; S_{j, d}^C
\end{equation}
We use the technique of~\cite{girshick15iccv} to backpropagate through the pooling layer, computing the subgradients of the loss with respect to each coordinate in the last convolutional feature map.
For each coordinate and channel in the ROI pooling output of a region, the gradients are passed to the convolutional feature map coordinate with the highest activations during the forward pass:
\begin{equation}
\frac{\partial \, \mathcal{L}}{\partial \, S_{j, d}^C} \, = \
\sum_r \sum\limits_{
i \; | \; \pi(i, d, r) \, = \, j
}
\frac{\partial \mathcal{L}}{\partial \, S_{i, d, r}^R}
\end{equation}

\mypar{Advantages.}
Our free-form region representations focus better on the region of interest, leading to purer
representations. Additionally, they solve a common problem with bounding boxes: when objects of two
classes occur in a part-container relationship (i.e. a bird in the sky), their free-form region
proposals degenerate to the same bounding box. Hence higher network layers will receive two identical feature vectors for two different
regions covering different classes. This leads to confusion between the two classes, both at
training and test time.

\mypar{Incorporating region context.}
Several works have shown that including local region context improves semantic segmentation~\cite{dai15cvpr,girshick14cvpr,hariharan15cvpr},
as many object classes appear in a characteristic context~(e.g. a lion is more likely to occur in the savanna than indoors).
We take into account region context by performing ROI pooling also on their bounding boxes using~\cite{girshick15iccv}.
Hence we combine the advantages of using context with the advantages of free-form region representations.

\begin{figure}[!ht]
\centering
\includegraphics[width=1.0\textwidth]{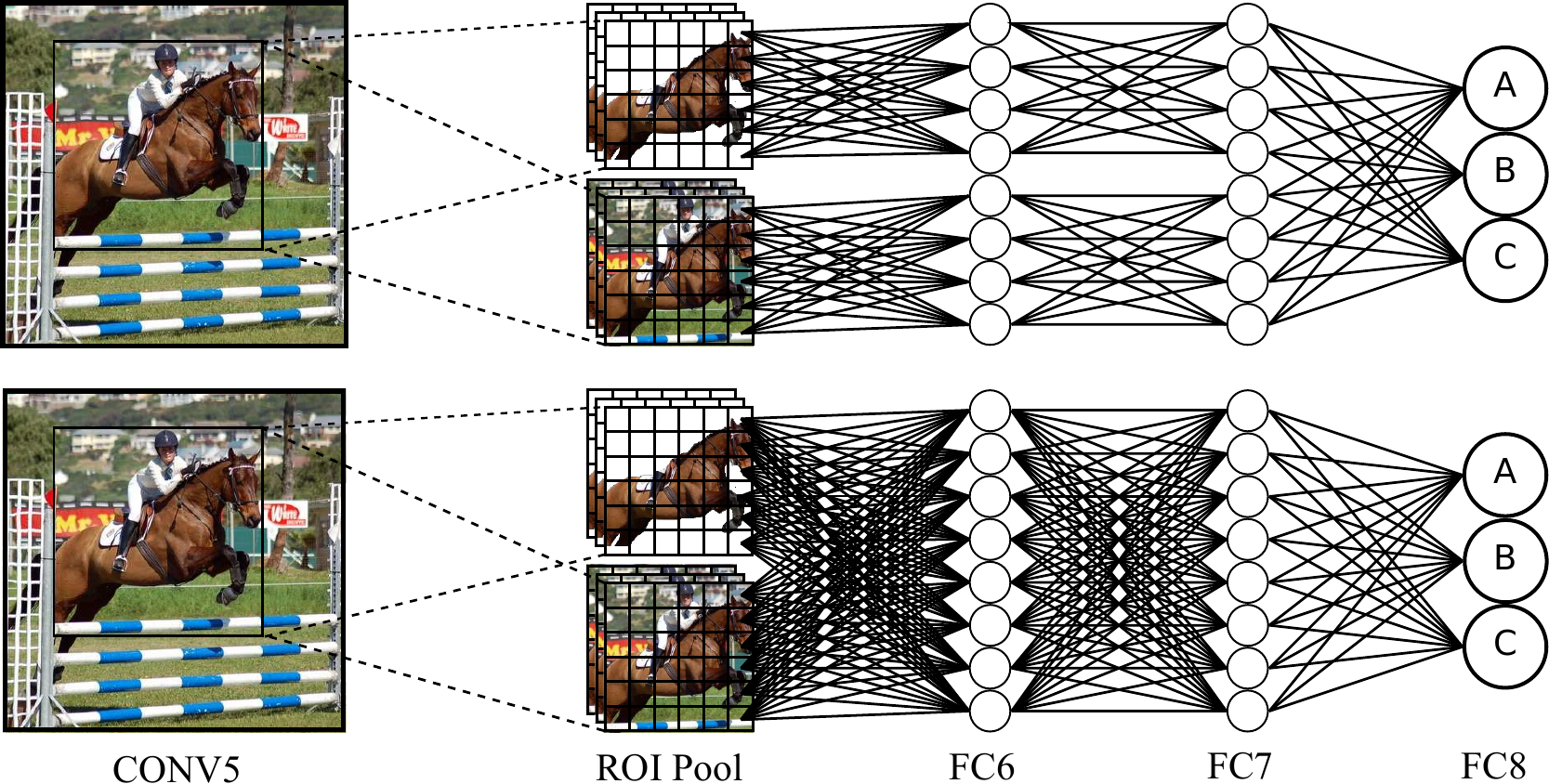}
\caption{
\textit{
We combine free-form region representations, which focus on the appearance of the region itself, with
bounding box based representations, which also capture context. We combine them using tied weights (above) and separate
weights (below).
}
}
\label{fig:cnn-context}
\end{figure}

As shown in Fig.~\ref{fig:cnn-context}, we combine region and bounding box representations using one of two
strategies:
\emph{(I) Tied weights.} We use the same fully connected layers with the same weights
for both region and bounding box representations and add the corresponding activations scores after the classification layers.
Hence the number of network parameters stays the same and the region and its context are handled identically.
\emph{(II) Separate weights.} We concatenate the representations of region and bounding box before applying the consecutive fully connected layers. 
This strategy roughly doubles the total number of weights of our overall network architecture, but can develop separate classifiers for each representation.

Since ROI pooling on bounding boxes and free-form regions are both differentiable, the combined representations are also differentiable and allow for end-to-end training.
We compare all representations experimentally (Table~\ref{tab:experiments-regioncontext}).

\mypar{Relation to~\cite{girshick15iccv,girshick14cvpr,dai15cvpr}.}
Girshick et al.~\cite{girshick15iccv} use a differentiable ROI pooling layer in Fast R-CNN for
bounding boxes only. 
Girshick et al.~\cite{girshick14cvpr} use free-form regions in R-CNN for semantic segmentation.
For each region proposal they set the color values of the background pixels to zero.
In our scheme we do not alter the image pixels of the input but pool exclusively over pixels inside the region.
Dai et al.~\cite{dai15cvpr} perform Convolutional Feature Masking on the last convolutional feature
map, followed by a Spatial Pyramid Pooling layer~\cite{he14eccv}, but did not backpropagate through
this layer.
Both~\cite{dai15cvpr,girshick14cvpr} combined free-form and bounding box representations.
Only~\cite{dai15cvpr} took representations after the convolutional layers, but their model was not able to perform backpropagation.
Both~\cite{dai15cvpr,girshick14cvpr} optimized for region classification instead of semantic segmentation.

\subsection{Attention to rare classes}
\label{sec:rareclasses}

Pixel-level class frequencies are often unbalanced~\cite{caesar15bmvc,mostajabi15cvpr,sharma14nips,sharma15cvpr,tighe13cvpr,yang14cvpr,eigen15iccv,farabet13pami,kekec14bmvc,byeon15cvpr,shuai15cvpr}.
This is typically addressed by using an inverse class frequency weighting $\frac{1}{P_c}$~\cite{mostajabi15cvpr,sharma14nips,eigen15iccv,farabet13pami}.
Since we have a pixel-level loss, we can simply plug this into
Eq.~\eqref{eq:loss-advanced}. However, we found that rare classes lead to large
weight updates resulting in exploding gradients and numerical problems. To avoid these issues, we
re-normalize the inverse frequency weights by a factor $Z$ so that the total sum of weights for each training image is $1$:
$\frac{1}{Z} \sum_c \frac{1}{P_c} \sum_{p=1}^{P} \; y_{p,c} = 1$.

\subsection{Efficient evaluation of the pixel-level loss}
\label{sec:efficienteval} 

Evaluating the loss for each pixel separately is computationally expensive and redundant, because
different pixels belonging to the same highest scoring region for a class are assigned the same
score $S_{r,c}$. Hence we partition the set of region proposals for a training image into a
set of non-overlapping, single-class regions using the ground truth. We then reformulate
Eq.~\eqref{eq:loss-advanced} into an equivalent loss in terms of these regions. This reduces the
cost of loss evaluation by a factor 1000.
\section{Experiments}
\label{sec:experiments}

\subsection{Setup}

\mypartop{Datasets.} We evaluate our method on two challenging datasets:
SIFT Flow~\cite{liu11pami} and PASCAL Context~\cite{mottaghi14cvpr}.
SIFT Flow contains 33
classes in 2688 images.  The dataset is known for its extreme class imbalance~\cite{liu11pami,farabet13pami,eigen15iccv}.
We use the provided fixed split into 2488 training images and 200 test images.

PASCAL Context provides complete pixel-level annotations for both things and stuff classes in the popular PASCAL VOC
2010~\cite{everingham15ijcv} dataset. It contains 4998 training and 5105 validation images.  As there
is no dedicated test set available, we use the validation images exclusively for testing.  We use
the 59 classes plus background commonly used in the literature~\cite{dai15cvpr,dai15iccv,long15cvpr,zheng15iccv}.

\mypar{Evaluation measures.}
Semantic segmentation methods typically measure global accuracy and class-average accuracy. Global accuracy is the percentage of correctly labeled pixels in the dataset. But since class frequencies typically follow a power-law
distribution, it is mostly influenced by a few common classes.
Class-average accuracy instead takes all classes into account equally and it is generally considered a better measure.
It first computes the accuracy for each class separately, and then averages over classes. Both measures are standard for SIFT Flow.
The most common evaluation measure on PASCAL Context is mean Intersection-over-Union~(IOU)~\cite{everingham15ijcv}.
For each class one divides the number of pixels of the intersection of the predicted and ground truth class by their union.
Then the average is taken over classes.

\mypar{Network.}
We use the state-of-the-art classification network VGG-16 \cite{simonyan15iclr}
pre-trained for image classification on ILSVRC 2012~\cite{russakovsky15ijcv}. We use the layers up to CONV5,
discarding all higher layers, as the basis of our network.
We then append a free-form ROI pooling layer~(Section~\ref{sec:freeform}),
a region-to-pixel layer, a softmax layer and pixel-level loss~(Section~\ref{sec:endtoend}, Fig.~\ref{fig:cnn-architectures}c).
To include local context, we combine region and entire bounding box using separate weights~(Section~\ref{sec:freeform}).

\mypar{Regions.}
We use Selective Search~\cite{uijlings13ijcv}, which delivers three sets of region proposals, one per color space~(RGB, HSV, LAB).
During training we change the set of region proposals in each mini-batch to have a more diverse set of proposals without the additional overhead of having three times as many regions.
We use region proposals with a minimum size of 100 pixels for SIFT Flow, and 400 pixels for PASCAL Context.
This results in an average of 370 proposals for SIFT Flow and 150 proposals for PASCAL Context, for each of the three color spaces.
Additionally we use all ground truth regions at training time.
This is especially important for very small objects that are not tightly covered by region proposals.

\mypar{Training.}
The network is trained using Stochastic Gradient Descent~(SGD) with momentum.
For 20 epochs we use a learning rate of 1e-3, followed by 10 epochs using learning rate 1e-4.
All other SGD hyperparameters are taken from Fast R-CNN~\cite{girshick15iccv}.
We use either an inverse-class frequency weighted loss~(referred to as {\em balanced} below) or a natural frequency weighted
loss~({\em unbalanced}).

\subsection{Main results}

\mypartop{SIFT Flow.}
We compare our method to other works on SIFT Flow test in Table~\ref{tab:experiments-siftflow}.
We first compare in the balanced setting, which takes rare classes into
account.
Hence we train our model for the loss described in Section~\ref{sec:rareclasses} and compare to methods using class-average accuracy.
We achieve $64.0\%$, which substantially outperforms the previous state-of-the-art, including the fully convolutional method~\cite{eigen15iccv} by $+8.3\%$ and the region-based method~\cite{caesar15bmvc} by $+8.4\%$.

We also compare in the unbalanced setting using global accuracy, which mainly measures performance on common classes.
Hence we train our model for the loss in Eq.~\eqref{eq:loss-advanced}.
This yields a competitive $84.3\%$ global accuracy, outperforming most previous methods, and coming close to the state-of-the-art~\cite{eigen15iccv} ($86.8\%$).

\begin{table}[t]
\centering
\small
\adjustbox{valign=t}{
\begin{minipage}[t]{.47\linewidth}
\raggedright
\begin{tabular}{L{2.1cm}l C{1cm} C{1cm} C{1cm}}
Method				&	Year	&	Class Acc.	&   Global Acc.	\\
\hline
\hline
Byeon~\cite{byeon15cvpr}	&	2015	&	22.6		&	68.7	\\
Gould~\cite{gould14eccv}	&	2014	&	25.7		&	78.4	\\
Tighe~\cite{tighe10eccv}	&	2010	&	29.4		&	76.9	\\
Pinheiro~\cite{pinheiro14icml}	&	2014	&	30.0		&	76.5	\\
Gatta~\cite{gatta14cvprw}	&	2014	&	32.1		&	78.7	\\
Singh~\cite{singh13cvpr}	&	2013	&	33.8		&	79.2	\\
Shuai~\cite{shuai15cvpr}	&	2015	&	39.7		&	80.1	\\
Tighe~\cite{tighe13cvpr}	&	2013	&	41.1		&	78.6	\\
Keke\c{c}~\cite{kekec14bmvc}	&	2014	&	45.8		&	70.4	\\
\\
\end{tabular}
\end{minipage}}
\hfill
\adjustbox{valign=t}{
\begin{minipage}[t]{.47\linewidth}
\raggedleft
\begin{tabular}{L{2.1cm} C{1cm} C{1cm} C{1cm}}
Method				&	Year	&	Class Acc.	&   Global Acc.	\\
\hline
\hline
Sharma~\cite{sharma14nips}	&	2014	&	48.0		&	79.6	\\
Yang~\cite{yang14cvpr}		&	2014	&	48.7		&	79.8	\\
George~\cite{george15cvpr}	&	2015	&	50.1		&	81.7	\\
Farabet~\cite{farabet13pami}	&	2013	&	50.8		&	78.5	\\
Long~\cite{long15cvpr}		&	2015	&	51.7		& \textbf{85.2}	\\
Sharma~\cite{sharma15cvpr}	&	2015	&	52.8		&	80.9	\\
Caesar~\cite{caesar15bmvc}	&	2015	&	55.6		&	-	\\
Eigen~\cite{eigen15iccv}	&	2015	&	55.7		& \textbf{86.8}	\\
Ours				&	2016	&  \textbf{64.0}	& \textbf{84.3}	\\
\\
\end{tabular}
\end{minipage}}

\caption{
\textit{
Evaluation on SIFT Flow test.
We show results for our model trained for either a balanced or an unbalanced loss.
We also show results of previous works, where we report the maximum result for each metric if multiple results are given.
}
}
\label{tab:experiments-siftflow}
\end{table}

\mypar{PASCAL Context.}
We also evaluate our method on the recent PASCAL Context dataset~\cite{mottaghi14cvpr}.
In Table~\ref{tab:experiments-pascalcontext} we show the results using either a balanced or an unbalanced loss.
Our balanced model achieves $49.9\%$ class-average accuracy, outperforming the only work that reports results for that measure~\cite{long15cvpr} by $+3.4\%$.
Our unbalanced model achieves competitive results on global accuracy~($62.4\%$) and reasonable results on mean IOU~($32.5\%$).

\begin{table}[t]
\centering
\begin{tabular}{l C{1.4cm} C{1.6cm} C{1.8cm} C{1.6cm}}
Method						&	Year	&	Class Acc.	&	Global Acc.	&	Mean IOU	\\
\hline
\hline
O2P~\cite{carreira12eccv} 			&	2012	&	-		&	-		&	18.1		\\
Dai et al.~\cite{dai15cvpr}			&	2015	&	-		&	-		&	34.4		\\
Long et al.~\cite{long15cvpr}			&	2015	&	46.5		&	\textbf{65.9}	&	35.1		\\
Dai et al.~\cite{dai15iccv}			&	2015	&	-		&	-		&	35.7		\\
Zheng et al.~\cite{zheng15iccv}			&	2015	&	-		&	-		&	\textbf{39.3}	\\
\hline
Dai et al. (add. boxes)~\cite{dai15iccv}	&	2015	&	-		&	-		&	40.5		\\ 
\hline 
Ours						&	2016	&	\textbf{49.9}	&	62.4		&	32.5		\\
\\
\end{tabular}
\caption{
\textit{
Evaluation on PASCAL Context validation.
We show results using a balanced and an unbalanced version of our method, as well as the current
state-of-the art, where we always report the maximum result for each metric.
O2P results are from the errata of Mottaghi et al.~\cite{mottaghi14cvpr}.
Dai et al.~\cite{dai15iccv} train using additional bounding box annotations.
}
}
\label{tab:experiments-pascalcontext}
\end{table}

\mypar{Qualitative analysis.}

\begin{figure}
\begin{minipage}{\linewidth}
\centering
\includegraphics[width=1.0\textwidth]{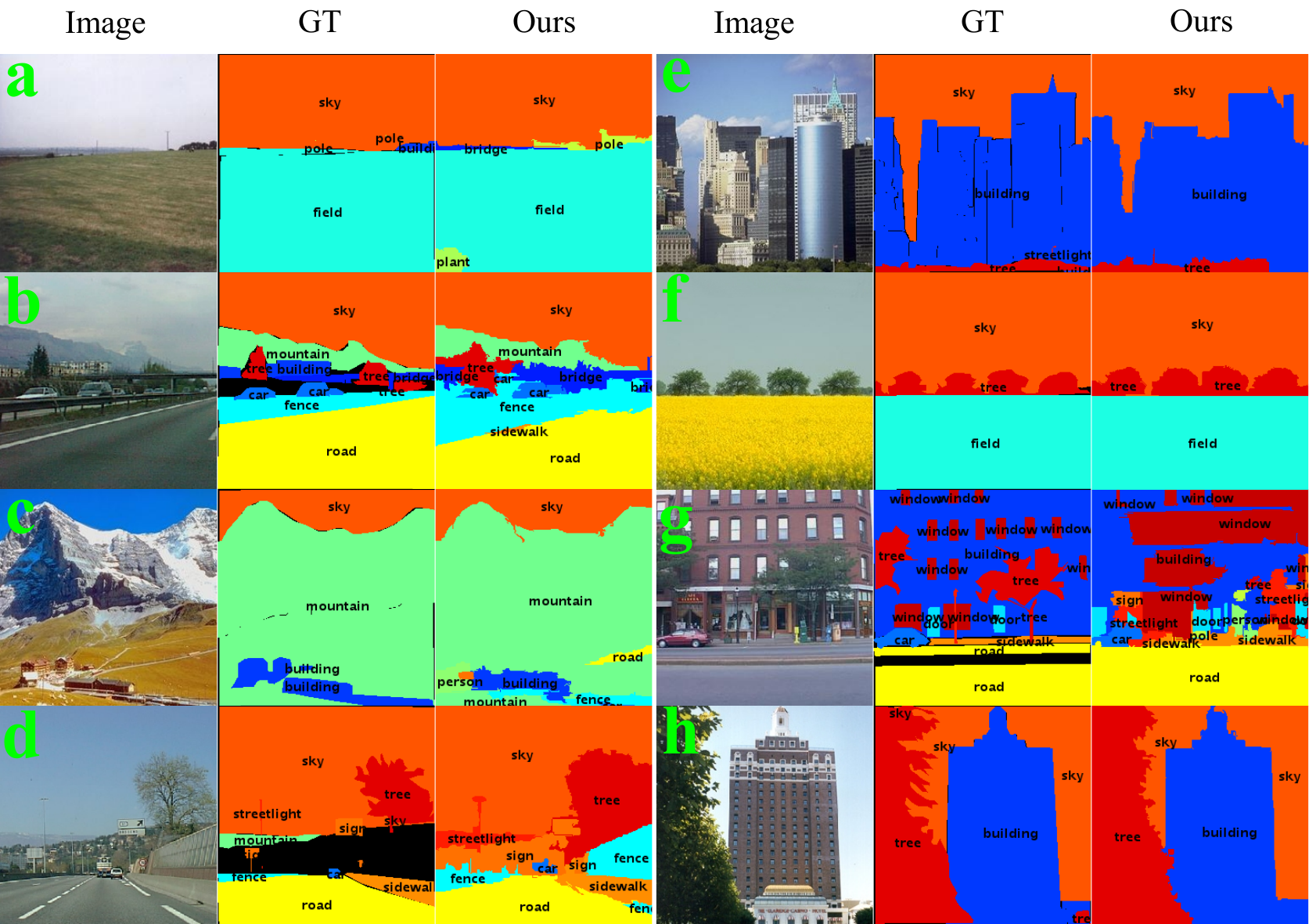}
\caption{
\textit{
Example labelings on SIFT Flow test.
We show an image, the ground truth labeling and the output of our balanced model.
}
}
\label{fig:examples-siftflow-balanced}
\end{minipage}

\begin{minipage}{\linewidth}
\centering
\includegraphics[width=1.0\textwidth]{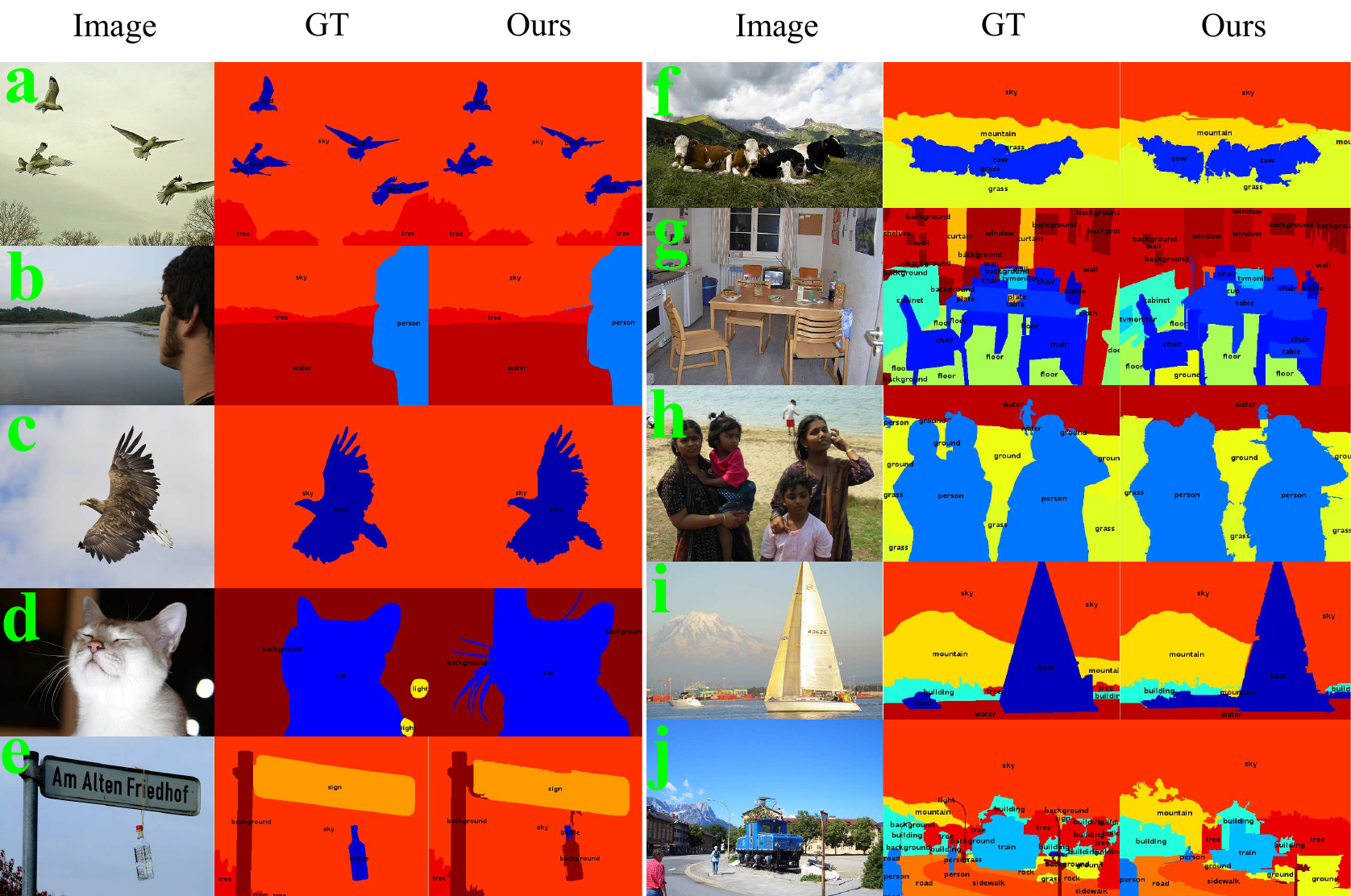}
\caption{
\textit{
Example labelings on PASCAL Context validation.
We show an image, the ground truth labeling and the output of our unbalanced model.
}
}
\label{fig:examples-pascalcontext}
\end{minipage}
\end{figure}

Fig.~\ref{fig:examples-siftflow-balanced} and~\ref{fig:examples-pascalcontext} show example
labelings generated by our method on SIFT Flow test and PASCAL Context validation.
Notice how our method accurately adheres to object boundaries, such as buildings~(Fig.~\ref{fig:examples-siftflow-balanced}e,~\ref{fig:examples-siftflow-balanced}h), birds~(Fig.~\ref{fig:examples-pascalcontext}a,~\ref{fig:examples-pascalcontext}c) and boat~(Fig.~\ref{fig:examples-pascalcontext}i).
This is one of the advantages of using a region-based approach.
Furthermore, our method correctly identifies small objects like pole (Fig.~\ref{fig:examples-siftflow-balanced}a) and the streetlight~(Fig.~\ref{fig:examples-siftflow-balanced}d).
This is facilitated by our method's ability to adaptively select the scale on which to do recognition.
Finally, notice that our method sometimes even correctly labels parts of the image missing in the ground truth, such as fence~(Fig.~\ref{fig:examples-siftflow-balanced}d) and cat whiskers~(Fig.~\ref{fig:examples-pascalcontext}d).

\subsection{Extra analysis}
\label{sec:extra-analysis}

\mypar{Accuracy at object boundaries.}
Following~\cite{kohli09ijcv,Krahenbuhl11nips},
we evaluate the performance on image pixels that are within 4 pixels of a ground truth object boundary.
We compare our method to the MatConvNet~\cite{vedaldi15mm} reimplementation
of Fully Convolutional Networks (FCN)~\cite{long15cvpr} in Table~\ref{tab:experiments-objectboundaries}.
On SIFT Flow test, FCN-16s obtains 37.9\% class-average accuracy on boundaries, while our method gets to 57.3\%.
When evaluated on all pixels in the image, FCN-16s brings 49.3\%, vs 64.0\% by our method.
Hence, our method is +19.4\% better on boundaries and +14.7\% on complete images.
Analogously, on PASCAL Context we get +4.9\% on boundaries and +1.8\% on complete images.
Since our improvements are consistently larger on object boundaries, we conclude that our method is especially good at capturing them, compared to the basic FCN architecture (Fig.~\ref{fig:cnn-architectures}a).

\begin{table}[t]
\adjustbox{valign=t}{
\begin{minipage}[t]{.46\linewidth}
\raggedright
\begin{tabular}{L{1.4cm} C{2.0cm} C{1.9cm}}
		& Boundaries	&	Full image	\\
\hline
\hline
FCN-16s		& 	\;\;37.9	&	\;\;49.3	\\
Ours		& 	\;\;57.3	&	\;\;64.0	\\
\emph{difference} &	\emph{+19.4}	&	\emph{+14.7}	\\
\hline
FCN-16s		&	\;\;34.0	&	\;\;48.1	\\
Ours		&	\;\;38.9	&	\;\;49.9	\\
\emph{difference} &	\emph{\,+4.9}	&	\emph{\,+1.8}	\\
\\
\end{tabular}

\end{minipage}}
\hfill
\adjustbox{valign=t}{
\begin{minipage}[t]{.50\linewidth}
\raggedleft
\begin{tabular}{L{2.0cm} C{2.4cm} C{1.5cm}}
    ROI pooling	& 		&	Class Acc.	\\
\hline
\hline
bounding box	& 			&	62.3		\\
region		& 			&	62.8		\\
region + box 	& tied weights		&	63.4		\\
region + box	& separate weights	&	64.0		\\
\hline
bounding box 	& purely rect. 	& 	59.3 		\\
\\
\end{tabular}

\end{minipage}}

\adjustbox{valign=t}{
\begin{minipage}[t]{.46\linewidth}
\raggedright
\caption{
\textit{
Class-average accuracy at object boundaries on SIFT Flow test~(top) and PASCAL Context validation~(bottom).
Improvements on boundaries are consistently larger than on full images.
}
}
\label{tab:experiments-objectboundaries}
\end{minipage}}
\hfill
\adjustbox{valign=t}{
\begin{minipage}[t]{.50\linewidth}
\raggedleft
\caption{
\textit{
Results on SIFT Flow test using free-form pooling, bounding box pooling or both.
We also report results when regions are rectangular even in the region-to-pixel layer (purely rectangular).
}
}
\label{tab:experiments-regioncontext}
\end{minipage}}
\end{table}

\mypar{End-to-end training.}
Our region-to-pixel layer enables end-to-end training of region-based semantic segmentation models.
We analyze how this end-to-end training influences performance, by comparing the baseline model
(Fig.~\ref{fig:cnn-architectures}b) to our model (Fig.~\ref{fig:cnn-architectures}c). To isolate the effect of end-to-end training, in
both models we perform ROI pooling on the bounding box only.
Hence all components of the two models are identical, apart from the region-to-pixel layer and the loss they are trained for.
On SIFT Flow test the baseline model achieves a global accuracy of $60.9\%$, compared to our $83.7\%$. 
We conclude that end-to-end training yields considerable accuracy gains over the baseline architecture in Fig.~\ref{fig:cnn-architectures}b.

\mypar{Softmax before max.}
Our application of the max before the softmax (Eq.~\ref{eq:model-advanced}) enables us to recognize each object at its appropriate scale~(Sec.~\ref{sec:endtoend}).
However, using the softmax before the max (Eq.~\ref{eq:model-simple}) yields an alternative model.
Interestingly, on SIFT Flow test our proposed order outperforms the alternative by $+8.7\%$ class-average accuracy.

\mypar{Importance of multi-scale regions.}
We argue that overlapping, multi-scale regions are important to unleash the full potential of region-based methods.
To show this, we train and test our model with non-overlapping regions~\cite{felzenszwalb04ijcv}.
This yields $60.0\%$ class-average accuracy on SIFT Flow test,
which is below the results when using multi-scale overlapping regions ($64.0\%$ class-average accuracy).

\mypar{Free-form versus bounding box representations.}

We analyze the influence of the different representations resulting from different ROI pooling
methods (Sec.~\ref{sec:freeform}). Keeping all else constant, we compare
(I) free-form ROI pooling, (II) bounding box ROI pooling, (III) their combination with tied weights
and (IV) their combination with separate weights. Results are shown in
Table~\ref{tab:experiments-regioncontext}.

Free-form representations perform $+0.5\%$ better than bounding box representations, demonstrating that
focusing accurately on the object is better.
Their combination does even better, yielding another $+0.6\%$ gain with tied weights (same number of model parameters) and $+0.6\%$ with separate weights. Hence both representations are complementary and best treated separately.

In all above experiments the region-to-pixel layer operates on free-form regions.
To verify the importance of the free-form regions themselves, we perform an extra experiment using purely rectangular regions (both in the region-to-pixel layer and during ROI pooling).
This lowers class-average accuracy by -4.7\%, demonstrating the value of free-form regions.
\section{Conclusion}

We propose a region-based semantic segmentation model with an accompanying end-to-end training
scheme based on a CNN architecture. This architecture combines the advantages of crisp object
boundaries and adaptive, multi-scale representations found in region-based methods with end-to-end
training directly optimized for semantic segmentation found in fully convolutional methods. We
achieve this by introducing a differentiable region-to-pixel layer and a differentiable free-form
ROI pooling layer. In terms of class-average pixel accuracy, our method outperforms the state-of-the-art
on two datasets, achieving $49.9\%$ on PASCAL Context and $64.0\%$ on SIFT Flow.

\mypar{Acknowledgements.} Work supported by the ERC Starting Grant VisCul.

\clearpage

\bibliographystyle{splncs-nobold}
\bibliography{bibtex/shortstrings,bibtex/calvin,bibtex/vggroup}
\end{document}